\documentclass[lettersize,journal]{IEEEtran}
\usepackage{amsmath,amsfonts}
\usepackage{algorithm}
\usepackage{array}
\usepackage[caption=false,font=normalsize,labelfont=sf,textfont=sf]{subfig}
\usepackage{textcomp}
\usepackage{stfloats}
\usepackage{url}
\usepackage{verbatim}
\usepackage{graphicx}
\usepackage{cite}
\usepackage{arydshln} 

\usepackage{booktabs}

\usepackage{multirow}

\usepackage{amssymb}
\usepackage{amsmath}
\usepackage{amsfonts}
\usepackage{algpseudocode,float}

\usepackage{xspace}
\usepackage{xcolor}
 
\makeatletter
\DeclareRobustCommand\onedot{\futurelet\@let@token\@onedot}
\def\@onedot{\ifx\@let@token.\else.\null\fi\xspace}

\def\ie{\emph{i.e}\onedot}

\def\etal{\emph{et al}\onedot}
\usepackage{tikz,xcolor,hyperref}

\definecolor{lime}{HTML}{A6CE39}
\DeclareRobustCommand{\orcidicon}{
\begin{tikzpicture}
\draw[lime, fill=lime] (0,0)
circle[radius=0.16]
node[white]{{\fontfamily{qag}\selectfont \tiny \.{I}D}}; 
\end{tikzpicture}
\hspace{-2mm}
}
\foreach \x in {A, ..., Z}{%
\expandafter\xdef\csname orcid\x\endcsname{\noexpand\href{https://orcid.org/\csname orcidauthor\x\endcsname}{\noexpand\orcidicon}}
}

\begin{document}

\title{Semantic Adversarial Attacks on Face Recognition through Significant Attributes}
\author{Yasmeen M. Khedr\orcidA{},~Yifeng Xiong\orcidB{}, and~Kun~He\orcidC{},~\IEEEmembership{Senior~Member,~IEEE}

\thanks{Manuscript received October 19, 2022.
}
\thanks{This work is supported by International Cooperation Foundation of Hubei Province, China (2021EHB011) and National Natural Science Foundation (62076105). Corresponding author: Kun He.}
\thanks{Yasmeen M. Khedr is with the School of Computer Science and Technology, Huazhong University of Science and Technology, Wuhan 430074, China, and also with the Faculty of Computers and Informatics, Zagazig University, Zagazig 44519, Egypt (e-mail: yasmeenkhedr@hust.edu.cn).}
\thanks{Yifeng Xiong and Kun He are with the School of Computer Science and Technology, Huazhong University of Science and Technology, Wuhan 430074, China (e-mail: xiongyf@hust.edu.cn; brooklet60@hust.edu.cn).}


}

\markboth{
}
{Khedr \MakeLowercase{\textit{et al.}}: Semantic Adversarial Attacks on Face Recognition through Significant Attributes}


\maketitle

\begin{abstract}

Face recognition is known to be vulnerable to adversarial face images. Existing works craft face adversarial images by indiscriminately changing a single attribute without being aware of the intrinsic attributes of the images. To this end, we propose a new Semantic Adversarial Attack called SAA-StarGAN that tampers with the significant facial attributes for each image. We predict the most significant attributes by applying the cosine similarity or probability score. The probability score method is based on training a Face Verification model for an attribute prediction task to obtain a class probability score for each attribute. The prediction process will help craft adversarial face images more easily and efficiently, as well as improve the adversarial transferability. Then, we change the most significant facial attributes, with either one or more of the facial attributes for impersonation and dodging attacks in white-box and black-box settings. Experimental results show that our method could generate diverse and realistic adversarial face images meanwhile avoid affecting  human perception of the face recognition. SAA-StarGAN achieves an 80.5\% attack success rate against black-box models, outperforming existing methods by 35.5\% under the impersonation attack. Concerning the black-box setting, SAA-StarGAN achieves high attack success rates on various models. The experiments confirm that predicting the most important attributes significantly affects the success of adversarial attacks in both white-box and black-box settings and could enhance the transferability of the crafted adversarial examples.
\end{abstract}


\section{Introduction}\label{sec1}

Face Recognition (FR)~\cite{DBLP:journals/corr/abs-2108-00401} is an important computer vision task widely used in solving authentication problems. FR can be categorized as Face Identification and Face Verification (FV). Over the past decades, FV, which determines whether a pair of face images belong to the same identity~\cite{Hou}, has achieved great achievements in various applications such as mobile payment, military, finance, surveillance security, and border control.

However, Szegedy \etal ~\cite{DBLP:journals/corr/SzegedyZSBEGF13} find that Deep Neural Networks (DNNs) are susceptible to adversarial examples. These adversarial examples have tiny perturbations added to the benign images that remain imperceptible to human vision but could mislead DNN models to produce incorrect predictions. Besides, some studies confirm the vulnerability of DNNs to input variations \cite{Hou,DBLP:journals/corr/SzegedyZSBEGF13,DBLP:journals/corr/GoodfellowSS14}. 
\IEEEpubidadjcol
Also, adversarial attacks can be divided into categories with different goals and assumptions on the attacker’s knowledge. White-box and black-box are two main settings based on the assumption of the attacker's knowledge. The former supposes that the attacker can access the model's parameter values, architecture, training method, inputs, outputs, and weights. Whereas the latter assumes that the attacker only has access to the inputs and outputs of the model but knows no information about the model \cite{DBLP:journals/corr/abs-2108-00401} 

There is a growing interest in adversarial studies for FR models \cite{DBLP:journals/mta/YangSW21,DBLP:conf/eccv/QiuXYYLL20,DBLP:conf/aaai/KakizakiY20,DBLP:conf/cvpr/DongSWLL0019}. Adversarial studies seek to generate adversarial face images to mislead facial recognition models. Methods used to manipulate the facial content includes face synthesis, identity swap, face morphing, face attribute manipulation, and expression swap~\cite{DeepFake}. 
Face attributes are among the emerging soft biometrics for modern security systems. Some studies use face attribute manipulation for different goals.  
Rozsa \etal~\cite{DBLP:conf/icpr/RozsaGRB16,DBLP:journals/prl/RozsaGRB19} 
propose the Fast Flipping Attribute technique
to mislead facial attribute recognition. 
Also, Mirjalili and Ross ~\cite{DBLP:conf/icb/MirjaliliR17} use the face attribute to modify the face image for a gender classifier. 
Recently, methods 
based on Generative Adversarial Network (GAN) have appeared that are used to manipulate facial attribute images, such as StarGAN \cite{DBLP:conf/cvpr/ChoiCKH0C18}, STGAN \cite{DBLP:conf/cvpr/0018DXLDZW19}, and AttGAN \cite{DBLP:journals/tip/HeZKSC19}.  
Joshi \etal~\cite{DBLP:conf/iccv/JoshiMSH19} use AttGAN to generate semantic attacks to deceive gender classifiers.
These studies are limited to classification problems instead of facial recognition.

Meanwhile, Qiu \etal~\cite{DBLP:conf/eccv/QiuXYYLL20} craft adversarial examples to mislead FR by changing the attribute individually and checking whether the generated image is adversarial until they find an adversarial example or failed after attempting the change. But, they craft an adversarial face image by indiscriminately distorting facial attributes without being aware of the significant facial attributes on each image. Hu \etal~\cite{AmT-GAN} propose Adversarial Makeup Transfer GAN (AMT-GAN) to generate adversarial face images, but it tends to produce high-quality images of females due to an imbalance of gender in the training dataset. 
These studies handle the attack on FR but they either suffer from the weakness of transferability to black-box models due to changing the face attribute randomly, or exhibit bias on genders due to the imbalance of data.

Our work aims to mislead FR models depending on changing the significant facial attributes. So, we propose a new attack method called the Semantic Adversarial Attack using StarGAN (SAA-StarGAN), which effectively and easily crafts semantic adversarial examples besides improving the attack transferability significantly by tampering with the significant facial attributes for each input image. These attributes are supposed to affect the decisions of different FV models, leading to deceiving the FV models and enhancing the adversarial transferability. 
In the white-box setting, we predict the most significant attributes for each input image by using either the cosine similarity (CS) or the probability score (PS) based on the Target Face Verification (TFV) model. Then, we change one or multiple via the StarGAN model in the feature space.  
The Attention Feature Fusion (AFF) method is used to fuse the features of inconsistent semantics to generate a realistic image and produce $\beta$ different values used for interpolation. In the black-box setting, SAA-StarGAN depends on predicting the most important attributes through the cosine similarity (CS) method. These attributes are modified on the input image according to their arrangement based on the prediction step by making an iterative loop to alter them sequentially until reaching the adversarial face images.

The empirical results confirm that predicting the most significant attributes (that will be changed first) plays an important role in successful attacks. Our SAA-StarGAN method outperforms other methods significantly on the attack success rate in the black-box setting and also preserves high attack success rates in the white-box setting for both impersonation and dodging attacks. Our method provides perceptually realistic images that maintain the source image identity to avoid confusing human perception. We also analyze the attention map of the TFV model that is attacked by our adversarial face images using gradient-weighted class activation. As a result, our method focuses on trivial features instead of prominent features. 

The main contributions of this work are summarized as follows:
\begin{quote}
\begin{itemize}
    \item We propose a novel attack method called Semantic Adversarial Attack using StarGAN (SAA-StarGAN) that enhances the transferability of adversarial face images by tampering with the critical facial attributes for each input image. 
    \item SAA-StarGAN generates semantic adversarial face images easily and effectively in white-box and black-box settings by predicting the most significant facial attributes using two techniques, cosine similarity or probability score, for impersonation and dodging attacks.
    \item We propose modifications on SAA-StarGAN to depend only on the output of the target model in a black-box setting by applying a linear search to find the optimal value of the interpolation coefficient that affects the generated face images. 
    \item The empirical results confirm that predicting the most significant attributes (which will be changed first) plays a vital role in a successful attack. Our SAA-StarGAN method outperforms other methods considerably on the white-box attack success rate and black-box adversarial transferability. Also, it provides perceptually realistic images that maintain the source image identity to avoid confusing human perception.
\end{itemize}
\end{quote}

\section{Related Work} 
\label{related}
In this section introduce the related work of generating adversarial examples for both image classification and FR models. 
\subsection{Adversarial Attacks on Images}

Many adversarial example generation methods have been proposed to mislead different image classification models. Most studies focus on generating restricted adversarial examples by adding perturbations to the input images. 
Szegedy \etal~\cite{DBLP:journals/corr/SzegedyZSBEGF13} first find the existence of adversarial examples for image classification, which transforms an image by a small amount to be undetectable and thereby changes how the image is classified. 
Goodfellow \etal~\cite{DBLP:journals/corr/GoodfellowSS14} propose a Fast Gradient Sign Method (FGSM) that uses the gradients of the neural network to generate adversarial examples. They just applied a one-step gradient update along the direction of the gradient sign at each pixel. 
Kurakin \etal~\cite{DBLP:conf/iclr/KurakinGB17a} propose a Basic Iterative Method (BIM), which applies FGSM perturbations of smaller magnitude for multiple iterations to improve the attack success rates. They clipped pixels in each iteration to avoid a large change 
Besides, Projected Gradient Descent (PGD) \cite{DBLP:conf/iclr/MadryMSTV18} is considered an extension of the BIM method to diversify the synthesized adversarial examples.

Dong \etal~\cite{DBLP:conf/cvpr/DongLPS0HL18} and Lin \etal~\cite{Lin-ICLR20} incorporate momentum into the iterative FGSM 
to boost the attack transferbility. 
Diverse methods have been proposed, such as  
designing an efficient saliency adversarial map for seeking the adversarial noise~\cite{DBLP:conf/eurosp/PapernotMJFCS16}, image denoising attack~\cite{DBLP:journals/tmm/ChengGJLFLL22},
and adding perturbation based on the attended regions and features~\cite{DBLP:journals/tmm/GaoHSYS22}.

Meanwhile, GAN is also used in the construction of adversarial examples due to its awesome ability to generate images.
Xiao \etal~\cite{DBLP:conf/ijcai/XiaoLZHLS18} propose AdvGAN, which constructs a generator network based on encoder-decoder to generate adversarial perturbation and then adds this perturbation to the original image to mislead the model. 
On the other hand,  Jandial \etal~\cite{DBLP:conf/iccvw/JandialMVB19} propose AdvGAN++, showing that latent features achieve higher attack success rates than AdvGAN and craft realistic images on CIFAR and MNIST datasets.
Song \etal ~\cite{DBLP:conf/nips/SongSKE18} propose a method for generating unrestricted adversarial examples based on the ACGAN from scratch instead of adding small perturbations on a source image for the classifier. Adversarial Transfer on Generative Adversarial Net (AT-GAN) \cite{ATGAN2019} 
generates non-constrained adversarial examples directly from any input noise, which aims to learn the distribution of the adversarial examples.

\subsection{Adversarial Attacks on Face Recognition} 
 
Recently, many adversarial attacks have been proposed for attacking the FR models, which can be divided into three categories: (1) adding adversarial perturbations, (2) manipulating facial attributes, and (3) physical attacks.

One line of study focuses on 
changing the facial appearance of input images by adding small perturbations in a specific region to be imperceptible to human eyes.  
Deb \etal~\cite{DBLP:conf/icb/DebZJ20} propose an automated adversarial face method called AdvFaces that generate minimal perturbations in the salient facial regions via GANs. 
Zhu \etal~\cite{DBLP:conf/icip/ZhuLC19} 
hide the attack information by the makeup effect to attack the eye regions only. 
Yang \etal~\cite{DBLP:journals/mta/YangSW21} propose Attentional Adversarial Attack Generative Network (A3GN) to deceive the FR model under impersonation attacks
, where attention modules and variational autoencoder are incorporated to learn semantic information from the target. 
Zhong and Deng \cite{DBLP:journals/tifs/ZhongD21} propose Dropout Face Attacking Network (DFANet) to improve the transferability by integrating the dropout in convolution layers in the iterative steps to generate adversarial examples. 
The second category is 
based on manipulating facial attributes. Rozsa \etal~\cite{DBLP:journals/prl/RozsaGRB19} propose Fast Flipping Attribute (FFA) technique which found that the robustness of DNNs against adversarial attacks varies highly between facial attributes.  Kakizaki and Yoshida \cite{DBLP:conf/aaai/KakizakiY20} focus on the serious risks resulting from the ineffectiveness of conventional certified defenses against adversarial examples that are not restricted to small perturbations. They use image translation techniques to generate unrestricted adversarial examples by translating the source image into any desired facial appearance with large perturbations.  Qiu \etal \cite{DBLP:conf/eccv/QiuXYYLL20} introduce SemanticAdv, which can generate unrestricted adversarial examples 
by altering a single facial attribute.

The physical attacks can be generated by a variety of different tools, such as adding some adversarial face accessories~\cite{DBLP:journals/tissec/SharifBBR19}, 
natural makeup \cite{NaturalMakeup} and adversarial patches~\cite{DBLP:journals/istr/RyuPC21}. These methods make the attacks more dangerous in the physical world. 

\section{Face Recognition Attack}
\label{FR_Attack}
In this section, we first provide the problem definition. 
Then we present a detailed description of our method 
in white-box and black-box settings.  Ultimately, we introduce a preliminary method called Random Selection
as a baseline method. 
\subsection{Problem Definition}
The main purpose of the FR model is to recognize the input image by training the model on dataset $D(x,y)$, where $x$ is a face image sampled according to the latent distribution, and $y$ is the corresponding ground-truth label. According to \(f(x):x\rightarrow{}y\), the model can predict the label for each input face image.
The main goal is to generate an adversarial face image $x_{adv}$ that is similar to the original image $x$ but misleads the FV model, \ie $FV(x_{adv}) = y' \neq y$. 

On the other hand, adversarial attacks are divided into two types: dodging and impersonation attacks \cite{DBLP:journals/corr/abs-2108-00401}. The dodging attack (untargeted attack) is developed to fool the target model such that the output is a random identity excluding the original one. In contrast, the impersonation attack (targeted attack) misleads the target model by recognizing the adversarial face image as a specified target identity. 
For the dodging attack, we generate $x_{adv}$, 
which is identified as not the same identity as $FV(x_{adv}) \neq y$. For impersonation attack, we seek to make the model recognize $x_{adv}$ as the same identity of another given image such that $FV(x_{adv}) = y_{tgt}$.
\subsection{Semantic Adversarial Attack (SAA-StarGAN)}
We propose an efficient attack method that first predicts the most significant facial attributes for each input image. Then, we generate $x_{adv}$ in the intermediate layers instead of the output layers because this increases internal feature distortion, leading to improved performance.
There are two steps in the proposed framework for \textit{a white-box attack}:
1) Predict the most significant attributes for each input image; 
2) Generate $x_{adv}$ by modifying one or multiple most significant attributes. 

These attributes are changed by using the StarGAN model. StarGAN model~\cite{DBLP:conf/cvpr/ChoiCKH0C18} consists of a single generator $G$ and a discriminator $D$ trained to learn to translate images from one domain to another. 

\subsubsection{Significant Attribute Prediction}
\begin{figure}[t]
\centering
\includegraphics[width=0.77\columnwidth]{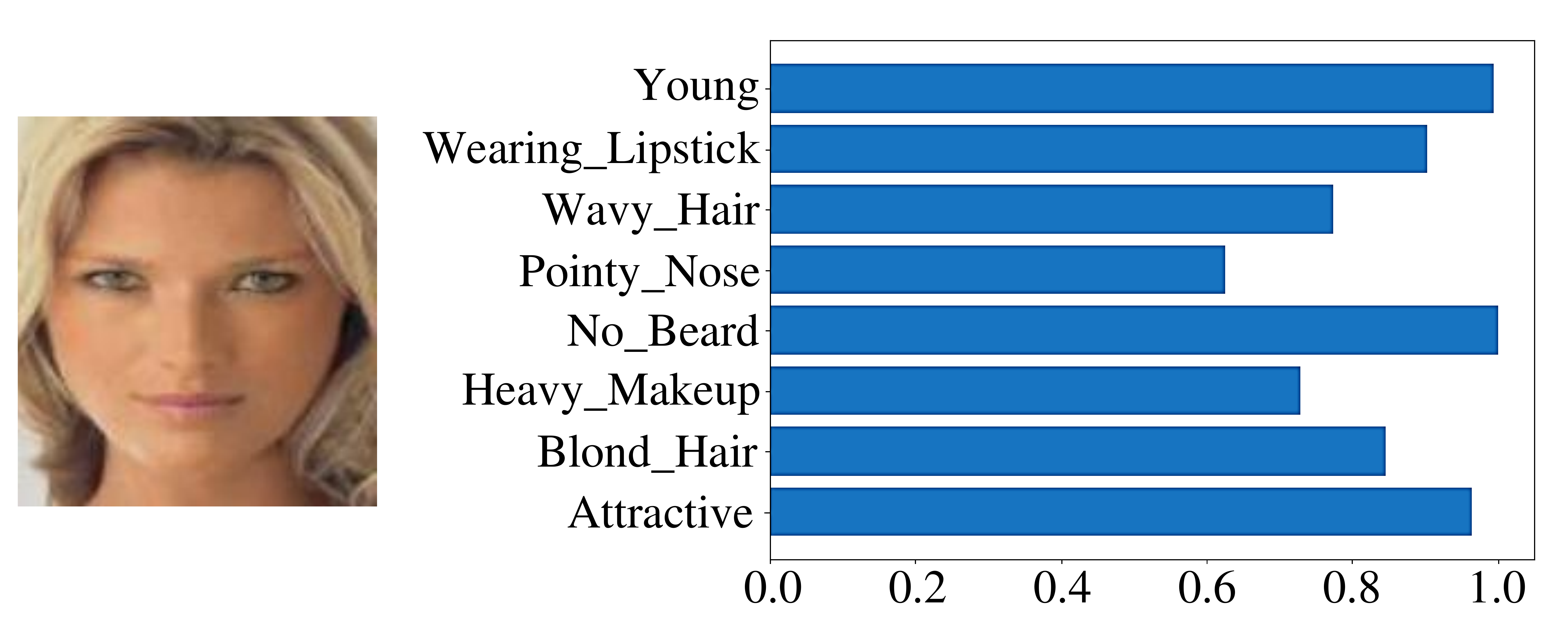} 
\caption{Output from the attribute prediction model using FaceNet.}
\label{probability}
\end{figure}

We propose using either the cosine similarity (CS) or the probability score (PS) to detect the significant attributes. The corresponding methods are denoted as SAA-StarGAN-CS and SAA-StarGAN-PS, respectively. 
We first apply CS or PS to predict the most significant attributes and compare their results. 
Consequently, we retrain the $G$ of StarGAN on all the facial attributes to use in the significant attribute prediction step.

a) \textbf{Cosine Similarity (CS)}: 

We utilize the TFV models 
\cite{DBLP:conf/cvpr/SchroffKP15, DBLP:conf/cvpr/DengGXZ19} to obtain the important attributes by computing the cosine similarity  between the output features of the TFV model. More details of the algorithm for the predicted most significant attributes using CS are shown in Supplementary.
For image attributes $a = (a_1, a_2, a_3,...,a_K)$ where $a_i$ indicates the $i^{th}$ attribute, and $K$ the total number of attributes,  
we first use StarGAN to change each $a_i$ of the input $x$ to get the image synthesis $x^*_{a_i}$.  
Then, we extract the synthesis image features $f_{x^*_{a_i }}$, and the original image features $f_x$ by the TFV model, 
and calculate their cosine similarity to get $S_{a_i}$, that  
refers to the degree of change in the output of TFV by changing attribute $a_i$. 
We then sort the attributes in $a$ in the ascending order according to the similarity score values to get the most important attributes $C = (c_1, c_2, c_3,...,c_K)$ where $c_i$ is the $i^{th}$ sorted attribute. The less the cosine similarity, the more the significance of the attribute. 
The cosine similarity, as illustrated in Eq. (\ref{cosine}), is a way to measure the similarity between two vectors, ranging from 0 to 1. 
 
\begin{equation} \label{cosine} 
S_{a_i} = CS(f_x, f_{x^*_{a_i }}) 
        = \frac{f_x \cdot f_{x^*_{a_i }}}{\|f_x \| \cdot \| f_{x^*_{a_i }}\|}.
\end{equation}

b) \textbf{Probability Score (PS)}: 

The basic idea of the probability score method is to use the TFV model as the \textit{Attribute Prediction model (Att-Pred)} to predict the important attributes of the input. For each attribute $a_i$ in input $x$, we use the Att-Pred model to obtain the class probability score (PS), as shown in Fig. \ref{probability}.
If $a'_i$ is found in the original image $x$, we remove it in the synthesis image $x^*$ and vice versa. Therefore, we use the StarGAN model to change $a'_i$ to get the $x^*_{a'_i}$. 

Moreover, each attribute in the input face image has a different impact on the final decision. Thus, we need to calculate the probability value $P_{a'_i}$ of $a'_i$ for $x$ and $x^*$ to get the degree of change.  $\Delta P_{a'_i }$ indicates the degree of change in the probability value in image $x$ before and after changing $a'_i$ to determine the most significant attributes for each image $x$.
\begin{equation} \label{DeltaP}
\Delta P_{a'_i } = P_{a'_i }(x)-P_{a'_i }(x^*_{a'_i}),
\end{equation}
where
\begin{gather*}
x = a_1 a_2...a_i...a_k,\\
x^*_{a'_i} = a_1 a_2...a'_i...a_k.
\end{gather*}
Here $x^*_{a'_i}$ indicates the generated image after changing attribute $a'_i$. 

Finally, we sort the attributes in $a$ in descending order using $\Delta P_{a'_i }$ to obtain the most significant attributes $C$ to help us craft $x_{adv}$. These attributes represent the best attack effect. The algorithm of this step is illustrated in Supplementary. 

\subsubsection{Adversarial Face Image Generation}

\begin{figure}[ht]
\centering
\includegraphics[width=0.9\columnwidth]{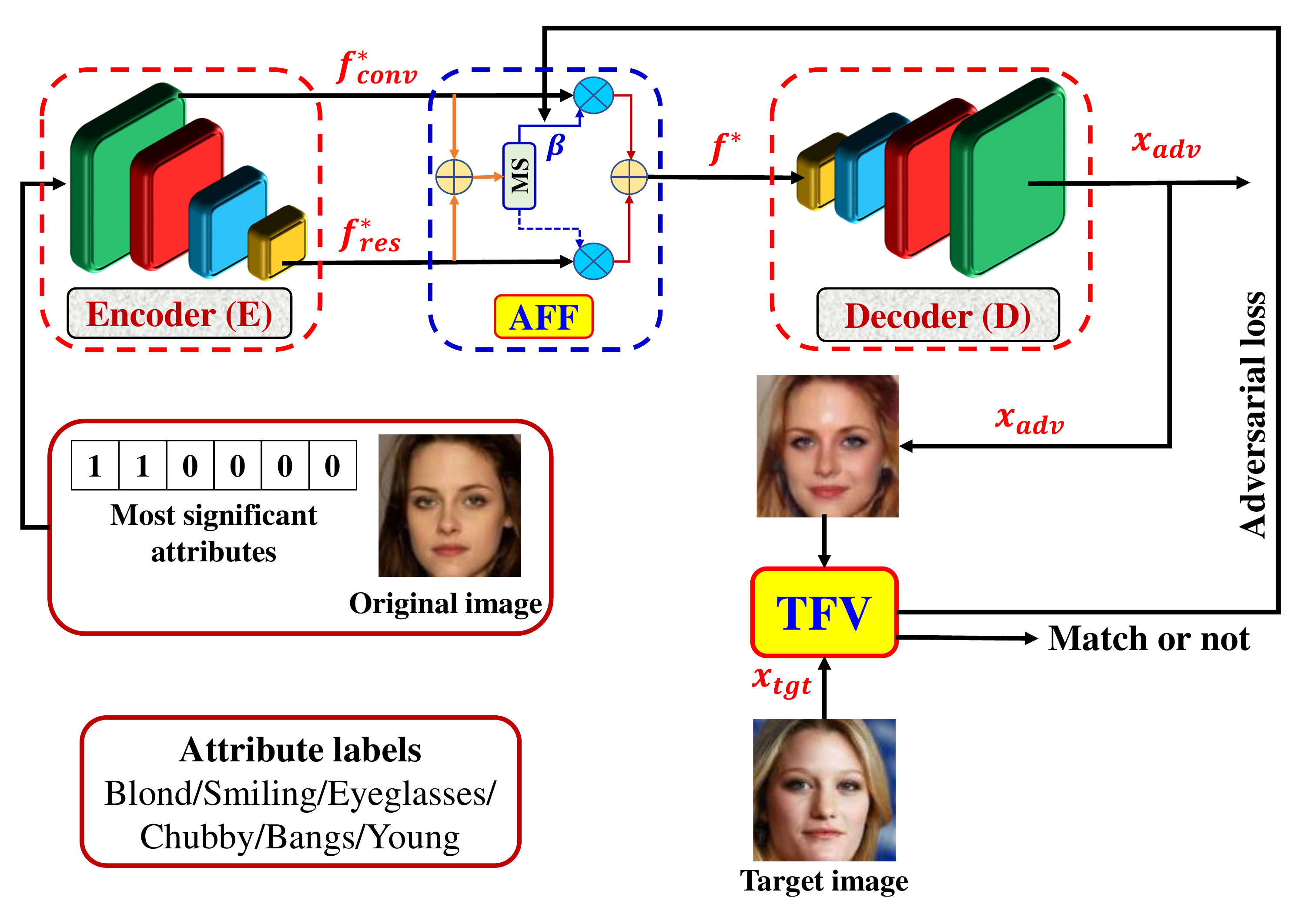}
\caption{The SAA-StarGAN attack framework. The original image and most significant attributes are fed into Encoder $(G_E)$ to change these attributes and then extract their features from different layers. The Attention Feature Fusion (AFF) framework is then used to generate $\beta$ and perform fusion between features. After that, the fused features are sent to the decoder $(G_D)$ to obtain the synthesis image. Finally, the $(TFV)$ model receives both the synthesis image and target identity to calculate the adversarial loss and optimize the $\beta$ value at the feature level.}
\label{framework}
\end{figure}

The second step focuses on generating the adversarial face images, as illustrated in Fig. \ref{framework}. 
To generate $x_{adv}$, we apply two kinds of perturbations using single or multiple attributes. 
SAA-StarGAN-CS-M and SAA-StarGAN-PS-M indicate the methods used for multiple attributes in CS and PS techniques, respectively.

a) \textit {For single attribute}, after completing the first step of predicting $C = (c_1, c_2, c_3,...,c_K)$, 
we use the generator $G$ of StarGAN, which we have already trained on facial attributes in the significant attribute prediction step. $G$ is composed of an encoder $(G_E)$ and a decoder $(G_D)$, as shown in Eq. (\ref{eq G}). The $(G_E)$ takes an input image $x$ and the single significant attribute $c_1$, then we get the output feature in intermediate layers. $(G_D)$ takes the feature as input, and outputs the synthesized image.

As illustrated in Fig. \ref{framework}, we use the $x$ as an input to the $G_E$ with significant attribute $c_1$. Then we extract the output features from the $conv$ layer $f^*_{conv}$ and the Residual Block layer $f^*_{res}$ of the encoder, as shown in Eqs.  (\ref{eq3}) and (\ref{eq4}).

\begin{equation} \label{eq G}
G = G_E  \cdotp  G_D,    \\
\end{equation}
\begin{equation} \label{eq3}
f^*_{conv} = G_E(x,c,conv\_layer),    \\
\end{equation}
\begin{equation} \label{eq4} 
f^*_{res} = G_E(x,c,res\_layer).\\
\end{equation}


On the other hand, we exploit the Attention Feature Fusion (AFF) method \cite{DBLP:conf/wacv/DaiGOWB21} to obtain the fused feature as an input to the decoder $G_D$.  Usually, addition and concatenation methods combine features. But recently, the attention mechanism has succeeded in many computer vision applications \cite{DBLP:journals/apin/LuH22}. So, we use AFF to combine the features. AFF is a framework that combines the features from different layers based on the Multi-Scale channel attention (MS) module  to overcome the inconsistent semantics among the input features and generate a more realistic image. More details about the AFF framework, including the MS module, are presented in \cite{DBLP:conf/wacv/DaiGOWB21}.
Eq. (\ref{eq alpha}) indicates the fusion weights $\beta$, where $\beta \in [0,1]$ is calculated from the attention weights generated by the MS module in AFF. 
As a result, we update the value of $\beta$ until the model is misleading. To get the better fused feature $f^*$, we apply an interpolation in the feature space, as shown in Eq. (\ref{eq fuse}). Finally, the decoder $G_D$ takes the fused feature $f^*$ as input and gets the synthesized face image $x^*$ as the output, as shown in Eq. (\ref{eq dec}).


\begin{equation} \label{eq alpha}
\beta = MS(f^*_{conv},f^*_{res} )      
\end{equation}
\begin{equation} \label{eq fuse} 
f^* = \beta \cdotp f^*_{conv}+(1-\beta) \cdotp f^*_{res}   
\end{equation}
\begin{equation} \label{eq dec} 
x^* = D(f^*) \\
\end{equation}

Our work obtains the $x_{adv}$ by modifying the significant attribute $c_1$ 
for each image through feature-level interpolation.
To achieve the impersonation attack, we use the $L_2$ loss function to minimize the distance between the face embedding of $x_{adv}$ and the target image $x_{tgt}$, as shown in Eq. (\ref{xadv}).

\begin{equation} 
\label{xadv} 
x_{adv} = Min_{x^*}  \parallel TFV(x^*) - TFV(x_{tgt} )\parallel_2^2
\end{equation}
The adversarial face image is crafted for dodging the attack by maximizing the distance between $x_{adv}$ and the $x$ in the feature space, as shown in the following:
\begin{equation} 
\label{xadv_dodging} 
x_{adv} = Max_{x^*}  \parallel TFV(x^*) - TFV(x)\parallel_2^2
\end{equation}

b) \textit {For multiple attributes}, similarly, we use the same $G$  of StarGAN, which has been used to modify a single attribute, but in this step, to change multiple significant attributes $C$ in the feature space.
In our experiments, we change two significant attributes, $c_1$ and $c_2$, represented as one-hot vectors. As a result, we obtain the synthesized image with a bigger change. After that, We study the effect of this change on attacking the face images. We do not change more than two attributes as we have tested using more significant attributes in $C$, but some of the resulting images are not realistic. 
\subsubsection{SAA-StarGAN in Black-Box Setting} 

A black-box attack differs from a white-box attack as the adversary has no access to the model’s gradient or parameters. Therefore, we propose modifications for our proposed method, SAA-StarGAN, to depend only on the target model’s output in the black-box setting. 

To generate a successful attack, we need two steps: 
1) Predicting the most significant attributes for the target model. This step is similar to that in the white-box setting, but we only use the CS method as it does not need to train a target model to predict the attributes. 
2) Doing a linear search to find the largest $\gamma$ value that affects the generated face image by changing the most significant attributes with $\gamma$ for each face image  in an iterative loop until the output misleads the model. 

More details of the algorithm for the black-box attack method are shown in Supplementary.
To change the most significant attributes $C$, we use $G_E$ from the StarGAN model to get the features. But, we make an iterative loop to alter the attributes according to the most important to the least until we meet the adversarial condition. We use a bilinear interpolation with a variable $\gamma$ value to generate the fused features $f^*_ {\gamma_i}$. 
Therefore, we make a linear search to find the optimal $\gamma_{opt}$ value according to the change in the confidence score. We create a vector $\gamma$ that consists of 100 random values drawn from [0, 1].  This vector is applied to study the effect of each value $\gamma$ in Eq. (\ref{eq fuse_black}). 

Then, the values are arranged according to the score change. 
For an impersonation attack, we select the optimal $\gamma_{opt}$ value that decreases the distance between the face embedding of the generated face image and the target face image. For the dodging attack, we select the optimal $\gamma_{opt}$ value that maximizes the distance between the face embedding of the generated face image and the input face image.  
Finally, we substitute the optimal $\gamma_{opt}$ value to generate the fused feature and feed it to a decoder to get the $x^*$ by the following equations:
\begin{equation} \label{eq fuse_black} 
f^* = \gamma_{opt} \cdotp f^*_1 +(1-\gamma_{opt}) \cdotp f^*_{c_ i} ,  
\end{equation}
\begin{equation} \label{eq dec_black} 
x^* = G_D(f^*). \\
\end{equation}

To guarantee that the generated face image will preserve semantic similarity from the original face image, we need to measure the semantic similarity $sim$ between the generated face image and the input face image to filter out the face images that are not realistic and control their quality. 
The adversarial face image is found when the adversarial criterion is achieved and the semantic similarity is above a threshold $th$ $(sim > th)$.

The above steps will repeat to add the next significant attribute in the ordered significant attributes until we find an adversarial example (Lines \ref{algorw18} - \ref{algorw34}).
\subsection{Random Attribute Selection based Attack}
\label{sec:randomSelectAttack}
This subsection presents a preliminary method 
, the Random Selection, as a baseline method to compare with our SAA-StarGAN method. We generate adversarial face images based on randomly selecting a set of attributes. These attributes are changed by using the StarGAN model. 

Therefore, we make two copies of the generator from StarGAN and re-train them separately using two different sets of attributes representing each attribute $a$ as a one-hot vector. The first generator, $G_1$, is trained by the first set $S_1$.
The second generator, $G_2$, is trained by the second set $S_2$
As a result, the overall Random Selection system involves three components, namely $G_1$, $G_2$, and the $TFV$ model. 

Firstly, \(G_1\) takes \(x\) with a specific attribute \(a_{s_1}\) from \(S_1\) to generate translated image with dimensions $H$, $W$, and $L$ for height, width, and channels, respectively. 
The second image synthesis \(x_2^*\) is then obtained using the translated image as the input to $G_2$ with another attribute \(a_{s_2}\)\ from \(S_2\), as shown in 
Eq. (\ref{random2}). Note that the two attributes are chosen randomly from the sets $S_1$ and $S_2$. Through permutations between the attributes from the previous sets, we change two attributes using the StarGAN model. 
 
Finally, interpolation is applied between the pair of images produced from \(G_1\) and \(G_2\) to generate \(x_{adv}\) as shown in Eq. (\ref{random3}). 
According to the presented procedure, 25 adversarial face images are generated for each $x$ due to the permutation between different attributes from the two sets.
\begin{equation} \label{random2} 
x_2^*= G_2 (G_1 (x, a_{s_1} ),a_{s_2}  )   
\end{equation}
\begin{equation} \label{random3} 
\begin{split}
x_{adv} = \alpha \cdotp x_1^*+(1-\alpha) \cdotp x_2^* \\  
\end{split}
\end{equation}

\section{Experimental Setup}
\label{exp_setup}
This section provide experimental setup, including dataset, models, baselines, evaluation metrics and implementation details of our method. 

\subsection{Dataset}
The proposed SAA-StarGAN method uses the CelebA dataset, which has 202,599 face images with 40 facial attributes and 10,177 identities \cite{DBLP:conf/iccv/LiuLWT15}, to generate semantic adversarial face images. This dataset is the most popular type used in the face recognition task. We use 40 facial attributes to train the StarGAN model in our work. Also, we randomly choose 5,000 different identities as the original images and 5,000 different identities as the target images. 
\subsection{Target Face Verification Models}
To evaluate the effectiveness of the proposed SAA-StarGAN, we choose ten state-of-the-art FV models, containing different model architectures and training loss functions. 
We use two of them as the white-box TFV models:  FaceNet ~\cite{DBLP:conf/cvpr/SchroffKP15} and ArcFace~\cite{DBLP:conf/cvpr/DengGXZ19}. 
The two models return 512-dimensional embeddings of the images. Besides, we use two other publicly available models, SphereFace~\cite{Sphereface} and CosFace~\cite{DBLP:conf/cvpr/WangWZJGZL018}, for evaluation. 
Then, we select different trained models under different backbones and loss functions such as ResNet-101 \cite{DBLP:journals/corr/RanjanCC17}, IResNet50 \cite{DBLP:conf/cvpr/HeZRS16,DBLP:conf/cvpr/DengGXZ19}, MobileFace \cite{Mobileface}, and ShuffleNet V2 \cite{DBLP:conf/eccv/MaZZS18} to demonstrate the effectiveness of our SAA-StarGAN on different models. We compute the optimal threshold $T$ based on the false-positive rate (FPR) for each FV model used in the evaluation. 
Details of each model are presented in Supplementary materials.
\subsection{Baselines} 
We adopt eight baseline methods to evaluate our attack method, including the Random Selection method presented in Section \ref{sec:randomSelectAttack}, the typical one-step attack method of FGSM~\cite{DBLP:journals/corr/GoodfellowSS14}, BIM~\cite{DBLP:conf/iclr/KurakinGB17a}, PGD~\cite{DBLP:conf/iclr/MadryMSTV18}, and MI-FGSM~\cite{DBLP:conf/cvpr/DongLPS0HL18}.
Besides, face attack methods such as  Sticker and Face mask attacks \cite{Sticker} are included. 
The facial attack method of SemanticAdv~\cite{DBLP:conf/eccv/QiuXYYLL20} is the most comparable method to ours, which modifies the face attribute to generate adversarial face images.
For the Random Selection method, five facial attributes are applied for \(G_1\): hair color (black-blond), heavy makeup, gender, and pale skin; while for \(G_2\), smiling, mouth slightly open, bangs, eyeglasses, and young attributes are applied. For FGSM, we set the perturbation as $\epsilon$ = 0.2 for \textit{$L_2$} attack and the number of iterations as 20 for BIM. Also, we set the perturbation as $\epsilon$ = 8 for PGD and MI-FGSM with pixel values in the range [0, 255]. The number of iterations is 40, and the decay factor $\mu$ = 1.0. 


\subsection{Evaluation Metrics} 
We use several evaluation matrices to estimate our attack effectiveness over different baselines. We select the attack success rate to evaluate the adversarial face images crafted by SAA-StarGAN.  To measure the similarity between images, we use the cosine similarity and use a threshold $T$ @ 0.1 \% FPR for each TFV model to decide whether the similarity between two identity persons matches. We can obtain cosine threshold $T_s$ from the Euclidean threshold after normalizing features as $T_s$ = $1-(T/2)$.

The attack success rate used for impersonation attack is computed as follows: 
\begin{equation} \label{eq12}   
success\_rate = \frac{(\# ImagePairs (x_{adv} , x_{tgt}) \geq T_s)}{(\# TotalImagePairs)},
\end{equation}
where ImagePairs consists of an adversarial face image generated by SAA-StarGAN and the matched target face.

The attack success rate used for dodging attack is computed as follows:
\begin{equation} \label{dodging}   
success\_rate = \frac{(\# ImagePairs (x_{adv} , x) < T_s)}{(\# TotalImagePairs)},
\end{equation}
where ImagePairs consists of an adversarial face image generated by SAA-StarGAN and the input face image.

Finally, we compute the \textit{Mean Square Error (MSE)}~\cite{DBLP:conf/cvpr/DongSWLL0019} and the \textit{Structural Similarity Index Measure (SSIM)} \cite{SSIM} to evaluate the quality between the original face images $x$ and the adversarial face images $x_{adv}$ for different attacks.
\subsection{Implementation Details} 
We use the Adam optimizer \cite{DBLP:journals/corr/KingmaB14}  with a fixed learning rate of 0.05 and up to 300 epochs for all experiments. SAA-StarGAN is implemented using PyTorch v1.7.0. For the PS, we train the TFV model for the attribute prediction task to have the class probability score (PS) for each attribute.
All experiments are conducted on a single Titan X GPU to generate adversarial face images.
We perform impersonation attacks based on the original and target images with different identities and dodging attacks based on the pairs of original images with the same identities. Firstly, all the selected face images pass through an MTCNN detector \cite{DBLP:journals/spl/ZhangZLQ16} to detect the face image and align images for the entire image. Then, we obtain the resized images in $112 \times  112 \times  3$, but inside each model it will carry out the specific input size due to the diversity of model inputs.

\section{Experimental Results }
\label{results}
To validate the effectiveness of SAA-StarGAN, we empirically evaluate our adversarial face images and illustrate that SAA-StarGAN achieves higher attack success rates against different FV models. On the other hand,  SAA-StarGAN boosts the adversarial transferability with high efficiency. The black-box setting demonstrates that SAA-StarGAN achieves high attack success rates significantly and shows the importance of 
significant attribute prediction. Besides, SAA-StarGAN can generate realistic and diverse adversarial face images.
Table \ref{transfer_on_FaceNet} and Table \ref{transfer_on_ArcFace} list the attack success rates of various methods, using FaceNet and ArcFace models to generate adversarial examples respectively, for impersonation and dodging attacks.

\subsection{Comparison for White-box Attacks}
To validate the efficacy of our attack method, we fisrt compare SAA-StarGAN with the baseline methods in the white-box setting. 
For a fair comparison, we train these baseline methods using FaceNet and ArcFace models on CelebA dataset with $T$ @ 0.1\% FPR for both impersonation and dodging attacks. Then we compute the attack success rate for each method. 
The first column in Table \ref{transfer_on_FaceNet} compares our SAA-StarGAN and the baselines in the white-box setting on the FaceNet model for the impersonation attack in (a) and dodging attack in (b). All variants of the proposed SAA-StarGAN method have reached a nearly 100 \% attack success rate, confirming that SAA-StarGAN can mislead the TFV models successfully. 
The second column in Table \ref{transfer_on_ArcFace} compares our SAA-StarGAN and the baselines in the white-box setting on the ArcFace model for the impersonation attack in (a) and dodging attack in (b).
For all the cases, it can be concluded that our SAA-StarGAN can effectively craft adversarial face images to fool TFV models in the white-box setting.


\subsection{Comparison for Black-box Attacks}
Most face recognition systems do not allow access to any internal information of the neural networks. In the black-box setting, the weights and network architectures are not included in the training process. So, it is essential to evaluate the vulnerability of FRs in both the transfer-based attack setting and the attack based on score confidence.
\subsubsection{Transferability Analysis}

The transferability across different models is one of the most important properties of adversarial examples. To demonstrate the transferability of the adversarial face images generated by SAA-StarGAN under two white-box models, we construct a dataset from successful adversaries crafted by the FaceNet and ArcFace models. Then we evaluate the attack success rate on nine different TFV models. 
We can observe from Table \ref{transfer_on_FaceNet} that for impersonation attacks, our SAA-StarGAN achieves a high attack success rate under different experimental conditions compared to various baselines. In contrast, the attack success rate of the Sticker and Face mask attacks has not exceeded 17 \%. FGSM and BIM methods exhibit the weakest transferability on all TFV models after the Sticker and Face mask attacks. The PGD, MI-FGSM, Random Selection, and SemanticAdv methods achieve better transferability than FGSM and BIM. The attack success rate of these attacks improves by 9.5 \% $\sim$ 38.6 \%, and our SAA-StarGAN outperforms them by a clear margin.

Although the Random Selection and SemanticAdv methods craft the adversaries by modifying the facial attributes, they only depend on changing random attributes or a single fixed attribute. We can see that the transferability of SAA-StarGAN-CS for the single attribute surpasses that of other methods in all cases. In addition, attacking the ShuffleNet V2 
model using adversarial face images generated by FaceNet for the SAA-StarGAN-CS achieves the best attack success rate of 54.30 \%, outperforming the baseline attacks by 15.7 \% $\sim$ 43.1 \%. 
Finally, we conclude that our SAA-StarGAN method outperforms others on all models besides maintaining a high attack success rate on the white-box setting. Similarly, our attack outperforms the other baselines significantly in dodging  attack, as presented in Table \ref{transfer_on_FaceNet}. We can observe that the adversarial face images generated by SAA-StarGAN-CS against the SphereFace exceed 14.1 \% $\sim$ 53.7 \% for the baseline attacks. Face mask based on the grid level is effective than PGD and MI-FGSM under dodging attacks. In addition, this attack is considered effective for transferability after our SAA-StarGAN.
\begin{table*}[]
\begin{center}
\caption{Transferability of the adversarial examples generated by SAA-StarGAN and the baselines against black-box models. We use adversarial examples generated on FaceNet model to attack nine different TFV models for impersonation \& dodging attacks. Values represent the attack success rate (\%). $^\ast$ indicates white-box attacks. The numbers marked in \textbf{bold} represent the best attack success rate. }
\label{transfer_on_FaceNet}%
\resizebox{\textwidth}{!}{%
\begin{tabular}{@{}lcccccccccc@{}}
\multicolumn{11}{@{}c@{}}{(a) Impersonation attack}\\\\
\toprule
Attack& FaceNet& ArcFace&ResNet-101& CosFace&SphereFace&  MobileFace&ShuffleNet V2 &IR50-Softmax&IR50-CosFace&IR50-SphereFace\\
\midrule
FGSM & 97.7$^\ast$&11.6&	02.6&	12.4&	18.4&	11.6&	13.8&	14.9&	10.1&	10.3\\
BIM& 98.0$^\ast$&	13.0&	08.6&	20.1&	21.9&	19.4&	19.6&	14.8&	09.8&	11.8  \\
PGD& 99.5$^\ast$&	12.5&	10.4&	24.6&	25.1&	27.5&	34.9&	17.4&	23.7&	20.7\\
MI-FGSM& 99.7$^\ast$&	13.4&	12.4&	29.1&	31.6&	28.3&	38.6&	21.7&	29.9&	21.8\\

SemanticAdv& 99.8$^\ast$&	32.4	&09.9&	22.1&	29.0&	28.1&	35.7&	16.2&	19.5&	14.7   \\
Random Selection& 99.7$^\ast$&	31.6&	09.5&	19.9&	25.1&	27.7&	31.4&	17.2&	29.5&	16.2   \\

Sticker& 100.0$^\ast$&	9.7&	7.4&	10.4&	11.5&	10.7&	11.2&	9.9&	10.5&	11.1\\
Face Mask& 95.4$^\ast$&	10.2&	9.6&	13.7&	17.0&	11.9&	15.7&	12.1&	11.4&	13.0\\
\midrule
SAA-StarGAN-CS& 100.0$^\ast$&\textbf{46.8}	&\textbf{31.5}&	\textbf{40.3}&	\textbf{53.0}&	\textbf{48.2}&	\textbf{54.3}&	\textbf{37.8}&	35.5&	\textbf{36.2}
\\ 

SAA-StarGAN-PS& 100.0$^\ast$&	45.5&	31.3&	39.6&	50.0	&48.1&	53.2&	34.5&	\textbf{36.5}&	35.3  \\ 
SAA-StarGAN-CS-M&	100.0$^\ast$&	44.3&	29.0&	39.9&	48.7&	46.7&	52.0&	34.9&	34.8&	36.0 \\ 
SAA-StarGAN-PS-M& 100.0$^\ast$&	46.2&	30.5&	40.2&	50.6&	47.0&	53.3&	34.3&	35.7&	34.4 \\
\bottomrule
\\
\multicolumn{11}{@{}c@{}}{(b) Dodging attack}\\\\
\toprule
  Attack& FaceNet& ArcFace&ResNet-101& CosFace&SphereFace&  MobileFace&ShuffleNet V2 &IR50-Softmax&IR50-CosFace&IR50-SphereFace\\
\midrule
FGSM & 94.6$^\ast$&	26.4&	14.4& 21.5&	30.2&	26.0&	28.3&	29.0&	21.0&21.4\\
BIM& 99.3$^\ast$&	28.0&19.5&	32.4&	33.7&	29.6&	33.2&	30.2&	20.7&	25.8  \\
PGD& 100.0$^\ast$&	50.0&	27.3&	62.4&	60.6&	49.9&	62.0&	41.5&	37.6&	42.9\\
MI-FGSM& 100.0$^\ast$&	51.1&	28.0&	60.3&	61.1&	50.1&	62.8&	43.2&	42.2&	48.4\\

SemanticAdv&  100.0$^\ast$ &	45.2&	20.4&	41.0&	39.5&	31.0&	48.1&	32.4&	33.1&	27.6  \\
Random Selection& 100.0 $^\ast$&	43.2&	20.0&	40.2&	36.2&	30.8&	42.9&	36.2&	47.2&	28.0  \\

Sticker& 100.0$^\ast$&	21.4&	9.8&	22.1&	23.5&	27.4&	26.4&	19.9&	21.4&	23.4\\

Face Mask&  100.0 $^\ast$&	55.6&	31.2&	63.5&	63.1&	52.1&	63.4&	40.1&	38.4&	43.7\\
\midrule
SAA-StarGAN-CS& 100.0$^\ast$&	\textbf{65.3}	&\textbf{44.6}&	\textbf{65.1}&	\textbf{77.2}&	\textbf{53.6}&	\textbf{76.8}&	\textbf{49.5}&	\textbf{53.4}&	\textbf{55.4}
\\ 
SAA-StarGAN-PS& 100.0 $^\ast$&	62.4&	43.5&	61.4&	71.8	&53.8&	69.4&		47.3&	49.5&	54.3  \\ 
SAA-StarGAN-CS-M&	100.0 $^\ast$& 51.5&	41.1&	61.2&	61.4&	54.9&	63.2&42.8&	47.6&	59.2 \\ 
SAA-StarGAN-PS-M& 100.0 $^\ast$&	54.7&		42.8&	64.4& 66.76&	54.5&	66.0&	47.2&	48.9&	51.0 \\
\bottomrule
\end{tabular}
}

\end{center}
\end{table*}
\begin{table*}[]
\begin{center}
\caption{Transferability of the adversarial examples generated by SAA-StarGAN and the baselines against black-box models. We use adversarial examples generated on ArcFace model to attack nine different TFV models for impersonation \& dodging attacks. Values represent the attack success rate (\%). $^\ast$ indicates white-box attacks. The numbers marked in \textbf{bold} represent the best attack success rate.}

\label{transfer_on_ArcFace}%
\resizebox{\textwidth}{!}{%
\begin{tabular}{@{}lcccccccccc@{}}
\multicolumn{11}{@{}c@{}}{(a) Impersonation attack}\\\\
\toprule
Attack& FaceNet& ArcFace&ResNet-101& CosFace&SphereFace&  MobileFace&ShuffleNet V2 &IR50-Softmax&IR50-CosFace&IR50-SphereFace\\
\midrule
FGSM & 20.1&	97.5$^\ast$&	12.4&	25.4&	23.4&	44.8&	41.4&	27.9&	33.0&	24.2\\
BIM& 21.8&	100.0$^\ast$&	16.4&	29.6&	27.1	&45.3	&44.6&	29.1&	35.5&	30.0\\
PGD &26.3&	99.9$^\ast$&	25.4&	42.5&	47.5&	53.8&	49.4&	45.3&	61.7&	45.1\\
MI-FGSM & 29.3&	100.0$^\ast$&	25.8&	48.9&	56.3&	65.8&	56.7&	51.9&	69.7&	52.7\\

SemanticAdv& 30.4&	95.8$^\ast$&	17.5&	32.6&	45.6&	51.5&	47.3&	37.5&	45.0	&30.7\\
Random Selection&30.6&	97.8$^\ast$&	19.4&	33.4&	38.9&	50.1&	48.7&	38.0&	49.5&	27.3\\

Sticker&18.7&	100.0$^\ast$&	8.9&	14.0&	10.1&	21.4&	20.9&	15.8&	12.4	&13.2\\
Face Mask&19.4&	100.0$^\ast$&	9.3&	16.5&	11.4&	22.6&	21.9&	17.0&	14.5&	13.9\\

\midrule
SAA-StarGAN-CS&\textbf{37.2}& 99.7$^\ast$	&\textbf{27.6}&	\textbf{54.2}&	\textbf{62.1}&	\textbf{74.3}&	\textbf{65.6}&	\textbf{65.9}& \textbf{80.5}&	\textbf{58.5}
\\ 

SAA-StarGAN-PS&36.7&99.8$^\ast$&	26.6&	52.0&	61.5&	73.5&	64.1&	64.7&	78.1&	57.5\\
SAA-StarGAN-CS-M&	34.2&99.1$^\ast$&	24.0&	53.0&	59.4&	67.1&	57.0&	61.9&	73.0&	52.9\\
SAA-StarGAN-PS-M&35.9&	99.2$^\ast$&	26.3&	53.7&	61.5&	68.5&	59.1&	64.4&	76.6&	56.4\\
\bottomrule
\\

\multicolumn{11}{@{}c@{}}{(b) Dodging attack}\\\\
\toprule
  Attack & FaceNet& ArcFace&ResNet-101& CosFace&SphereFace&  MobileFace&ShuffleNet V2 &IR50-Softmax&IR50-CosFace&IR50-SphereFace\\
\midrule
FGSM &34.1&	99.0$^\ast$&	32.7&	46.7&	44.7&	66.1&	62.7&	49.2&	54.3&	45.5\\
BIM&35.6&	100.0$^\ast$&	35.8&	51.5&	49.0&	67.2&	66.5&	51.0&	57.4&	49.9\\

PGD& 47.7&	99.9$^\ast$&	39.2&	63.2&	70.6&	72.6&	66.9&	64.9&	67.7&	55.5\\
MI-FGSM&48.2&	100.0$^\ast$&	41.5&	66.0&	73.0&	75.8&	70.0&	66.8&	69.8&	56.7\\

SemanticAdv&44.4&	99.9$^\ast$&	37.4&	53.2&	64.6&	80.5&	76.3&	66.5&	64.0&	52.7\\
Random Selection&45.6&	99.8$^\ast$&	38.9&	54.8&	60.3&	81.5&	77.1&	69.4&	61.9&	53.6\\

Sticker&21.3&	100.0$^\ast$&	25.4&	27.8&	30.1&	40.2&	38.4&	24.7&	26.8&	27.4\\

Face Mask&45.0&	100.0$^\ast$&	42.4&	64.7&	70.0&	74.5&	70.5&	68.7&	69.1&	58.4\\

\midrule
SAA-StarGAN-CS&\textbf{53.2}&	100.0$^\ast$&	\textbf{48.0}&	\textbf{75.6}&	\textbf{86.4}&	\textbf{94.6}&	\textbf{89.1}&	\textbf{77.1}&	\textbf{89.3}&	\textbf{69.9}\\
SAA-StarGAN-PS&52.4&	100.0$^\ast$&	47.2&	73.3&	83.8&	93.9&	88.7&	76.0&	88.1&	64.8\\
SAA-StarGAN-CS-M&49.2&	100.0$^\ast$&	44.6&	72.4&	80.4&	91.1&	86.0&	70.9&	72.1&	61.9\\
SAA-StarGAN-PS-M&50.9&	100.0$^\ast$&	46.6&	72.7&	81.5&	92.5&	87.1&	73.4&	75.6&	65.4\\

\bottomrule
\end{tabular}
}

\end{center}
\end{table*}

\begin{figure}[h]
\begin{center}
\centering
\includegraphics[width=0.50\textwidth]{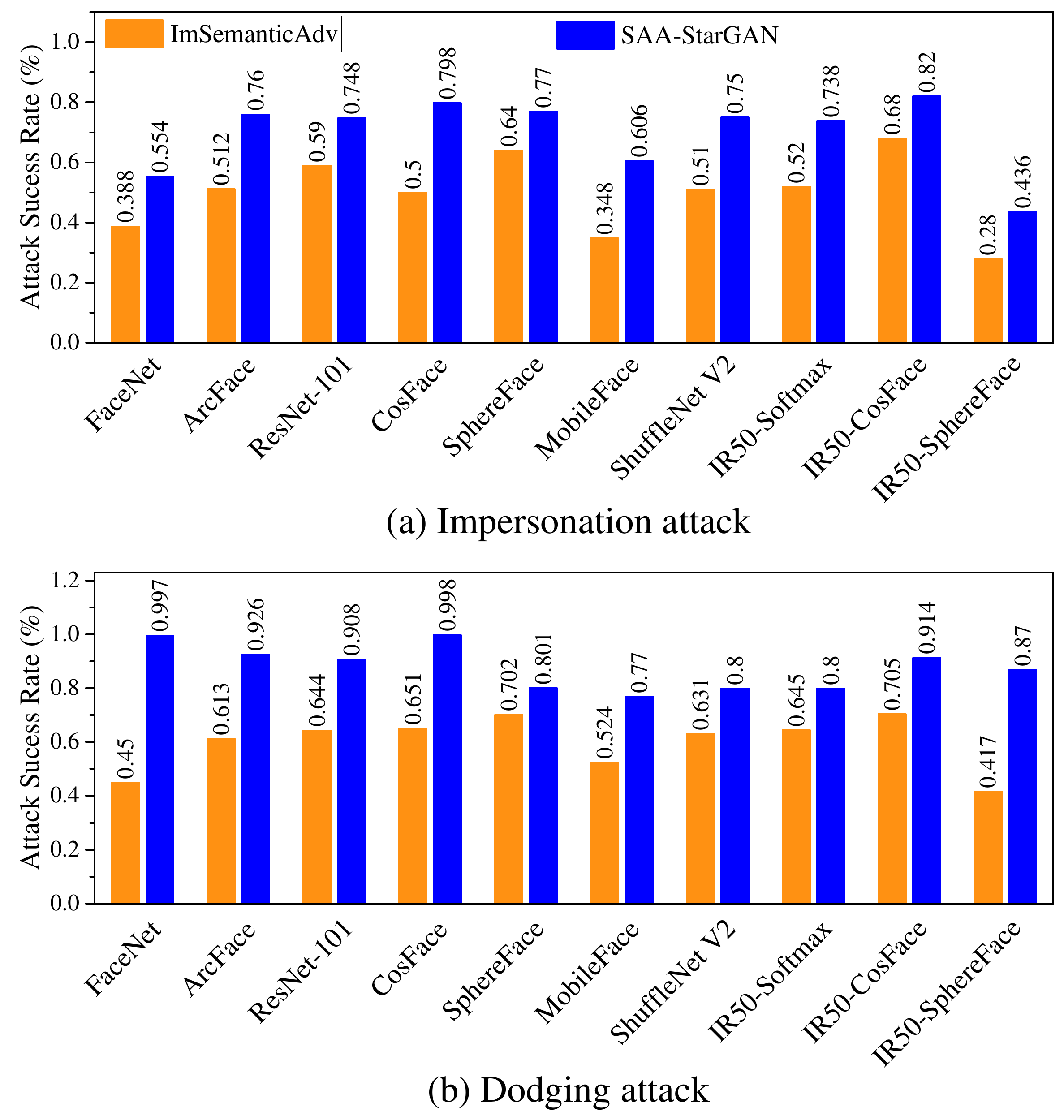}
\caption{The average attack success rate (\%) for black-box attacks.
SAA-StarGAN outperforms ImSemanticAdv by a clear margin.}
\label{black_box_bars}
\end{center}

\end{figure}

\begin{figure}[h]
\centering
\includegraphics[width=0.85\columnwidth]{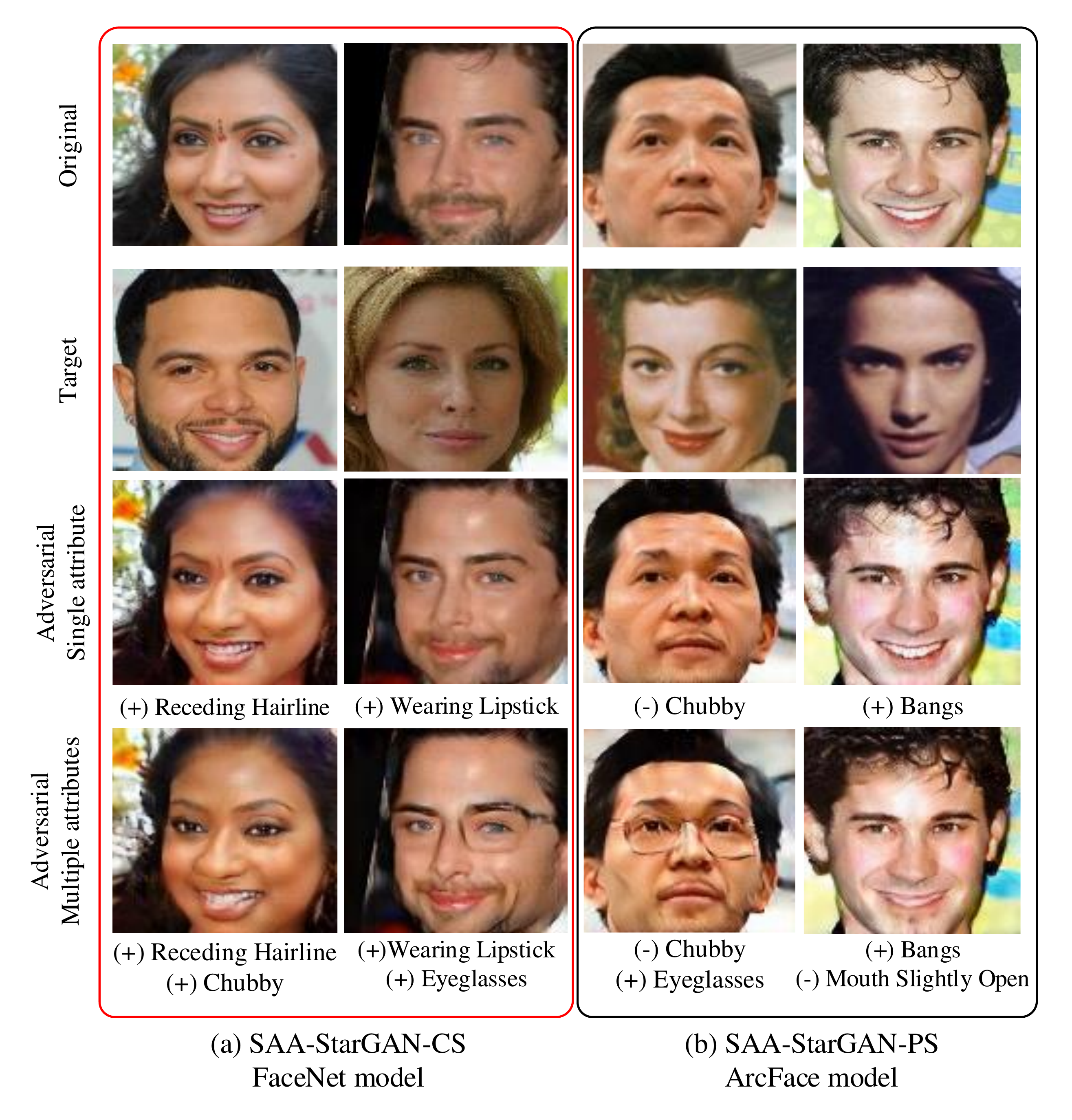}
\vspace{-0.5em}
\caption{Adversarial face images generated by SAA-StarGAN in  white-box setting. It shows the success of SAA-StarGAN in producing realistic images.  }
\label{fig: visualImages}

\end{figure}

\begin{figure}[h]
\centering
\includegraphics[width=0.95\columnwidth]{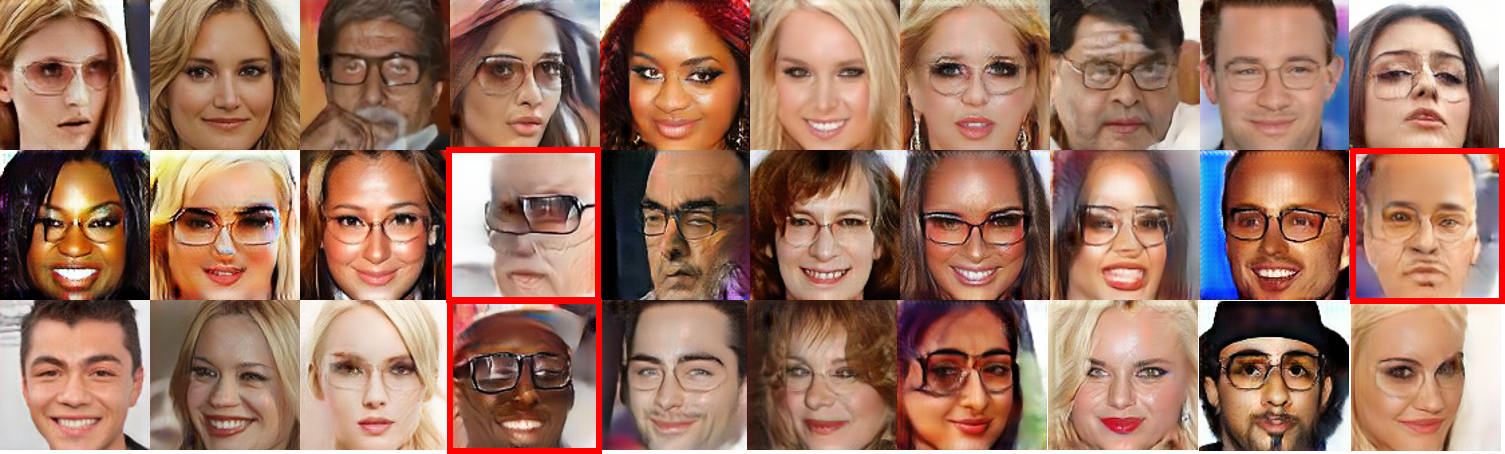} %
\caption{Adversarial face images generated by SAA-StarGAN in the black-box setting based on the score confidence. The figure illustrates the realistic face images, and the red borders indicate a few unrealistic adversarial face images generated by SAA-StarGAN. 
}
\label{SAA_black}
\vspace{1em}
\end{figure}
\begin{figure}[h]
\centering
\includegraphics[width=0.95\columnwidth]{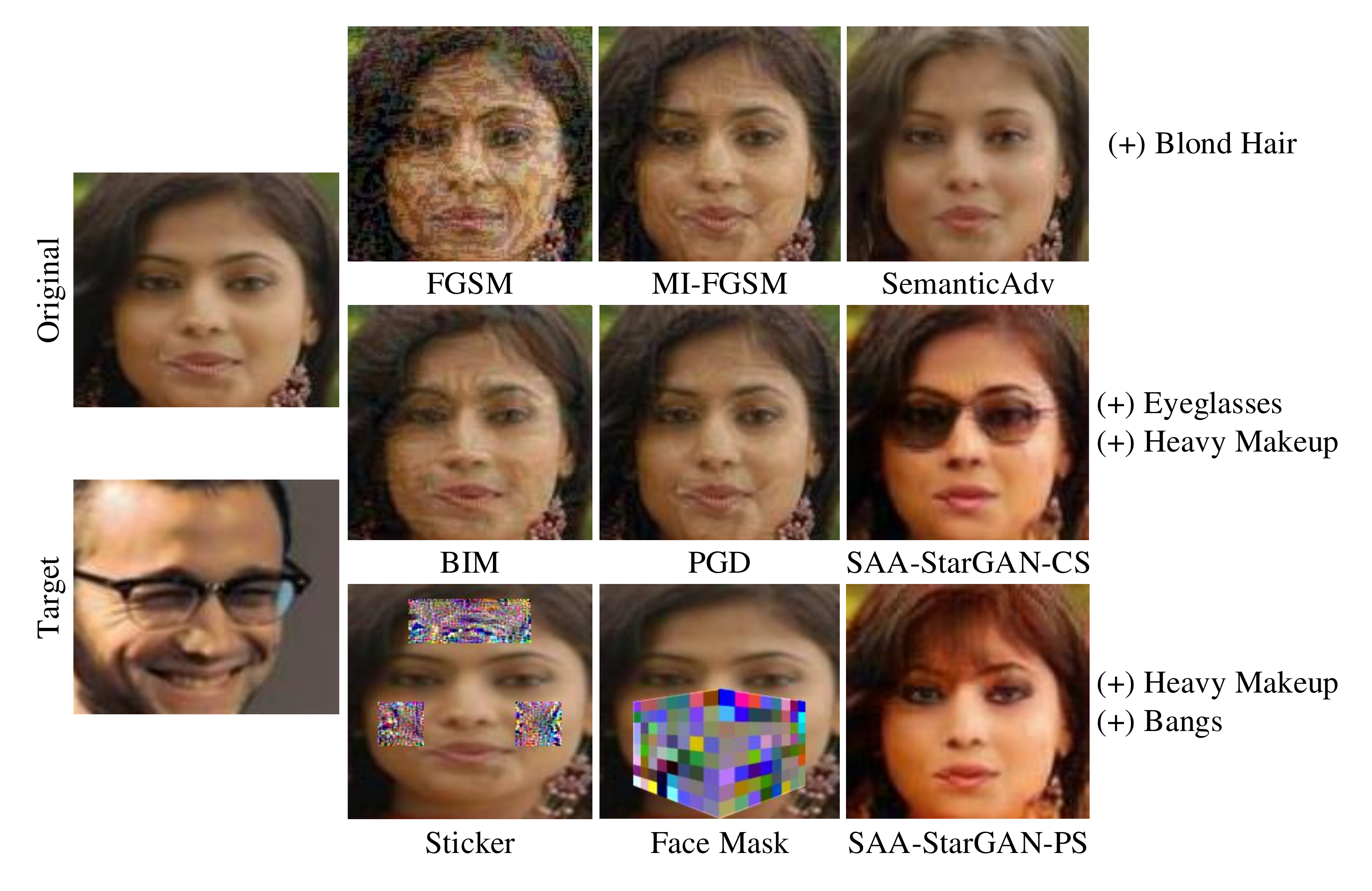} %
\caption{The adversarial face images generated on FaceNet by different methods. SAA-StarGAN can generate high quality adversarial face images as compared with the baselines.
}
\vspace{1em}
\label{compare}
\end{figure}
Table \ref{transfer_on_ArcFace} illustrates that our SAA-StarGAN significantly improves the transferability of adversarial face images crafted by the ArcFace model against different models over the baseline attacks.
The main reason is that SAA-StarGAN depends on manipulating the most important attributes that affect the decision. As clear from Table \ref{transfer_on_ArcFace},
SAA-StarGAN-CS outperforms the baselines of MI-FGSM, PGD, Random Selection, SemanticAdv, BIM, FGSM, Face Mask, and Sticker by 10.8, 18.8, 31.0, 35.5, 45.0, 47.5, 66.0, and 68.1 \% respectively under the impersonation attack, against IR50-CosFace model. To dodge an attack, The results are illustrated in Table \ref{transfer_on_ArcFace}.
We conclude that the main advantage of SAA-StarGAN-CS is that it does not need to train the TFV models for the attribute prediction tasks. We can directly apply different TFV models easily and efficiently.
Besides, we observe that the models within the same backbone have good transferability. So, the generated adversarial face images crafted on the ArcFace model against IR50-Softmax, IR50-CosFace, and IR50-SphereFace have high transferability. And ShuffleNet V2 and MobileFace have light weights which are easily attacked by the adversarial face images generated on ArcFace. In contrast, the FaceNet model is challenging to transfer to other models, where it is trained on IncepetionResNetV2 based on the softmax loss. 

We provide an additional experiment described in the Supplementary to confirm the effectiveness of the  SAA-StarGAN.
\subsubsection{Comparison of Attacks based on Score Confidence}

We also evaluate the performance of SAA-StarGAN of black-box attacks based on score confidence 
to empirically demonstrate that the proposed method in the black-box setting can generate adversarial images that mislead different FV models. We have improved the SemanticAdv method (denoted as ImSemanticAdv) to add multiple attributes in black-box setting, based on the random attribute selection. We compare SAA-StarGAN with ImSemanticAdv that follows our method's procedure when changing attributes. This method is performed by making a loop to change the random attributes until reaching the adversary randomly. The main results of black-box attacks based on score confidence for the impersonation and dodging are shown in Fig. \ref{black_box_bars}. We can observe that SAA-StarGAN outperforms ImSemanticAdv across all different models by a large margin for the impersonation and dodging black-box attacks. Compared with ImSemanticAdv, SAA-StarGAN improves the average attack success rate by 13 \% $\sim$ 29.8 \%, and 9.9 \% $\sim$ 54.7 \% under impersonation and dodging attacks, respectively.
Therefore, predicting the most important attributes of the target model affects the attack's performance significantly. 

\subsection{Visualization for SAA-StarGAN} 
The goal of adversarial face images is to mislead the FV models
Without fooling humans. Consequently, we illustrate samples of adversarial face images generated by SAA-StarGAN-CS and SAA-StarGAN-PS on FaceNet and ArcFace respectively in the white-box setting. As illustrated in Fig. \ref{fig: visualImages},  
the label of images, sign (+) indicates the addition of attribute while, sign (-) denotes removing the attribute. We can see that the proposed method achieves the goal of producing realistic images and the change on the images is slight.
 
 Fig. \ref{SAA_black} presents a set of adversarial face images that are generated on FaceNet in the black-box setting based on the score confidence. We observe that a few images are not realistic. But in general, the proposed method can generate realistic and diverse images.
 The comparison between SAA-StarGAN and baselines on FaceNet model is illustrated in Fig. \ref{compare}. Although two attributes are applied to the proposed SAA-StarGAN-CS and SAA-StarGAN-PS methods, the realistic of the generated adversarial face images are close enough to those produced by SemanticAdv with single attribute. On the other hand, the adversarial face images generated by FGSM, BIM, PGD, and MI-FGSM are not clear compared to our SAA-StarGAN. The face images generated by Sticker and Face Mask attacks are unrealistic and misleading to the human eye. 
We can conclude that the goal has been successfully achieved in the current study, as the presented faces are realistic and can be easily identified.

\subsection{Visualizing Attention on Adversarial Face images}
\begin{figure}[h]
\centering
\includegraphics[width=0.49 \textwidth]{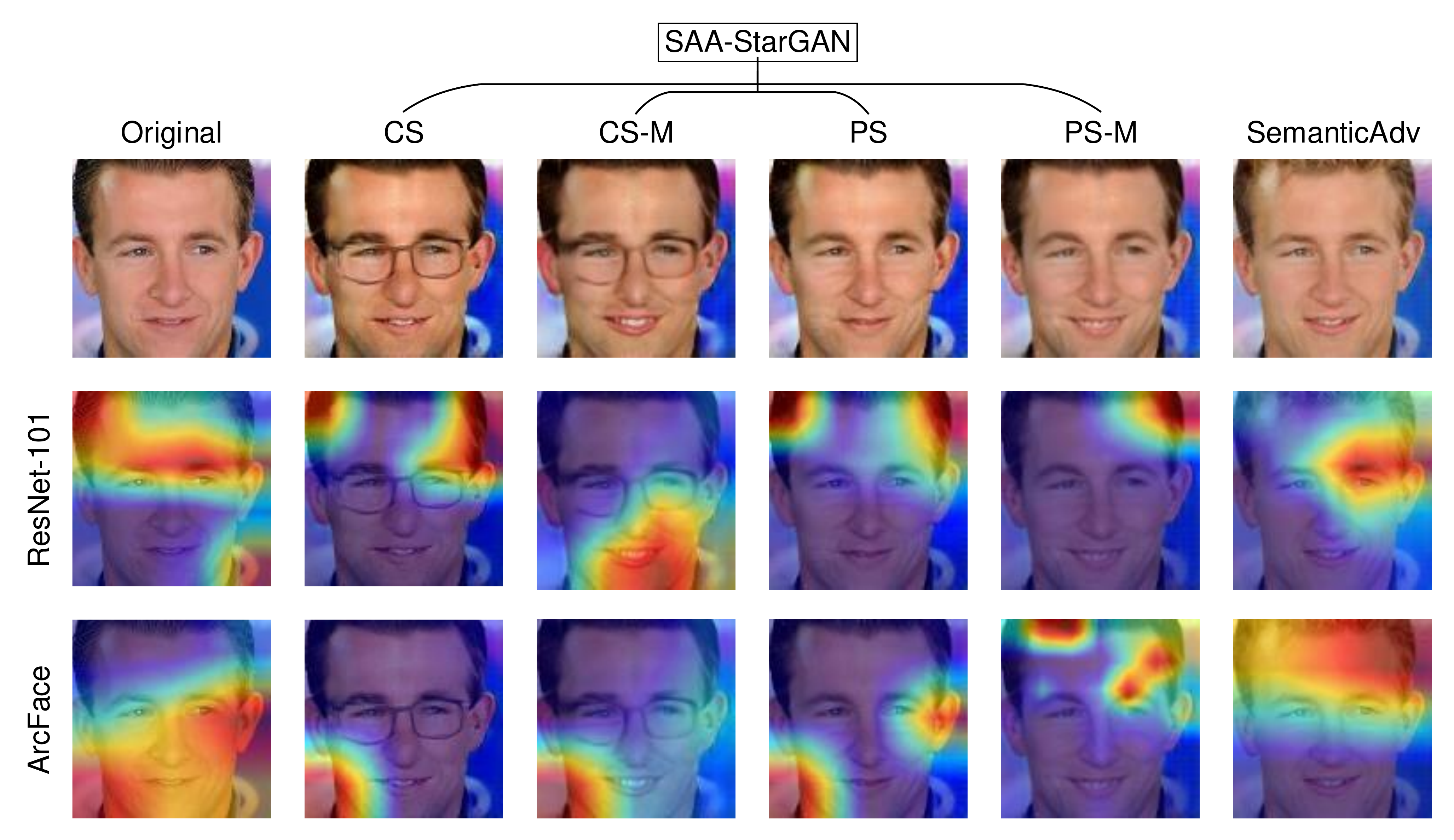} 
\vspace{0.5em}
\caption{Grad-CAM attention visualization for ResNet-101 and ArcFace models being attacked by adversarial face images are generated on the FaceNet model. This Figure compares our SAA-StarGAN based on the important facial attributes and SemanticAdv based on indiscriminate attributes.
}
\label{Grad_cam}

\end{figure}

In this subsection, we apply the Gradient-weighted Class Activation Mapping (Grad-CAM) \cite{gradcam}, an attention visualization technique, to find the discriminative regions in the image according to the TFV model. We use it to show the attention of the ArcFace and ResNet-101 models for both the original and the generated adversarial face images on the FaceNet model. In Fig. \ref{Grad_cam},
we can see that our SAA-StarGAN focuses on the trivial regions instead of the prominent regions, as opposed to the SemanticAdv method focuses on the most prominent regions in the face image. 
Thus, SAA-StarGAN  could improve the transferability significantly.
We conclude that the prediction step for the most significant attributes plays a key role in improving the attack transferability.
\subsection{Discussion on the Similarity to the Original Image}
It is evident in the literature that most methods succeed in generating adversarial face images. Therefore, we aim to measure the quality between the adversarial face and original images through MSE and SSIM. MSE measures the absolute errors, 
So the lowest value is the best. 
SSIM is used for measuring the similarity between two images by predicting the perceived quality of images. So, the highest value is the best.
We compare the adversarial face images crafted by our SAA-StarGAN
with the baseline methods. We select 2,000 original images and 2,000 adversarial face images for different attack methods for the evaluation. After that, we calculated the MSE and SSIM for each method separately, as shown in Table \ref{quality}. SAA-StarGAN-PS and SAA-StarGAN-CS for single attribute show the largest values of SSIM compared to others. Besides, they have the lowest values of MSE, which give them preference and make them more suitable for generating adversarial face images. The main reason is that our method depends on changing a single most significant attribute. 
This change leads to a slight modification in the adversarial face image, but with a high effect in misleading the different models.
On the other hand, SemanticAdv has a high structural similarity value. 
The Random Selection method randomly applies two attributes, which is lower than the SemanticAdv method in the structural similarity. 
In contrast, Sticker and Face Mask methods on FaceNet or ArcFace model show the highest absolute error. These methods mainly depend on pixel-level and grid-level, which add large perturbations to the face image, covering approximately 20 $\sim$ 30 \% area of the face image. We can see that any adversarial face images generated by FGSM, BIM, PGD, and MI-FGSM are expected to be perceptible to human eyes.
Generally, SAA-StarGAN-PS and SAA-StarGAN-CS for a single attribute are most similar to the original images and have the lowest error. Therefore, it is recommended to use our proposed method to craft the adversarial face images. 

\begin{table}[h]
\begin{center}
\caption{MSE and SSIM to compare the original and adversarial face images generated on FaceNet and ArcFace models by our SAA-StarGAN and the different baselines. Best values appear in bold.}\label{quality}
\scalebox{0.85}{
\begin{tabular}{lcccc}
\toprule
Attacks& \multicolumn{2}{@{}c@{}}{FaceNet} & \multicolumn{2}{@{}c@{}}{ArcFace} \\

\addlinespace[2pt]
\cline {2-3} \cline {4-5}
\addlinespace[2pt]
&MSE ($\downarrow$) &SSIM ($\uparrow$) &MSE ($\downarrow$) &SSIM($\uparrow$)\\
\midrule
FGSM & 0.034&	0.388&	0.036&	0.374	\\
BIM & 0.031&	0.596&	0.032&	0.585 \\
PGD& 0.026&	0.715&	0.026&	0.700\\
MI-FGSM& 0.028&	0.718&	0.029	&0.705\\
SemanticAdv& 0.023&	0.813&	0.023&	0.813  \\
Random Selection&0.034&	0.727&	0.034&	0.726\\
Sticker& 0.958&	0.009&	0.959&	0.008\\
Face Mask& 0.968&	0.007&	0.969&	0.006\\

\midrule

SAA-StarGAN-CS&0.022&	0.822&	0.020&	0.821	\\
SAA-StarGAN-PS&	 \textbf{0.015}$^\ast$ &	\textbf{0.854}&	\textbf{0.016}$^\ast$&	\textbf{0.846}\\
SAA-StarGAN-CS-M&0.040&	0.729&	0.036&	0.735	\\
SAA-StarGAN-PS-M&0.026&	0.792&	0.026&	0.784\\
\bottomrule
\end{tabular}}
\end{center}
\end{table}

\section{Conclusion}
\label{conclusion}

In this work, we present a new adversarial attack method called the SAA-StarGAN to generate semantic adversarial face images in both white-box and black-box settings. This method focuses on predicting the important facial attributes for each input image to change a few of the most important attributes to make the model promote the trivial features in the face images. 
In the white-box setting, SAA-StarGAN first predicts the significant attributes of each input image via TFV and uses the cosine similarity or probability score to determine the most significant attributes. Then the most important features are changed using a StarGAN model in the feature space. In the black-box setting, we make a loop to change the ordered significant attributes until the output is an adversary. Finally, the generated adversarial face images achieve good impersonation and dodging attacks. We observe that SAA-StarGAN could generate high-quality and realistic images through the experiments and achieve high attack success rates in the white-box and black-box settings. The proposed SAA-StarGAN exhibits significantly higher transferability than the state-of-the-art face attack methods. The results demonstrate the significant improvements in attacking various face recognition models due to the step of important attributes prediction for each input image.

\bibliography{Reference}
\bibliographystyle{IEEEtran}



\begin{IEEEbiography}[{\includegraphics[width=1in,height=1.25in,clip,keepaspectratio]{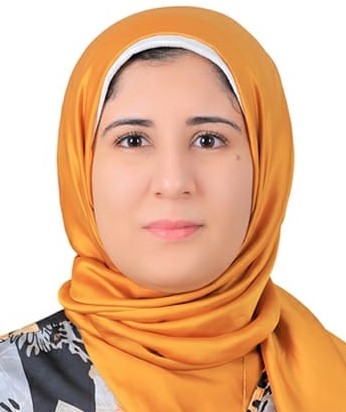}}]{Yasmeen M. Khedr} 
is currently pursuing the Ph.D. degree with Huazhong University of Science and Technology, Wuhan, China. She received the B.S. and M.S. degrees in information technology from Zagazig University, Egypt, in 2011 and 2017, respectively.
She is an Assistant Lecturer with the Department of Information Technology, Zagazig University. Her research interests include adversarial learning, artificial intelligence, multimedia security, image processing and computer vision.
\end{IEEEbiography}

\begin{IEEEbiography}
[{\includegraphics[width=1in,height=1.25in,clip]
{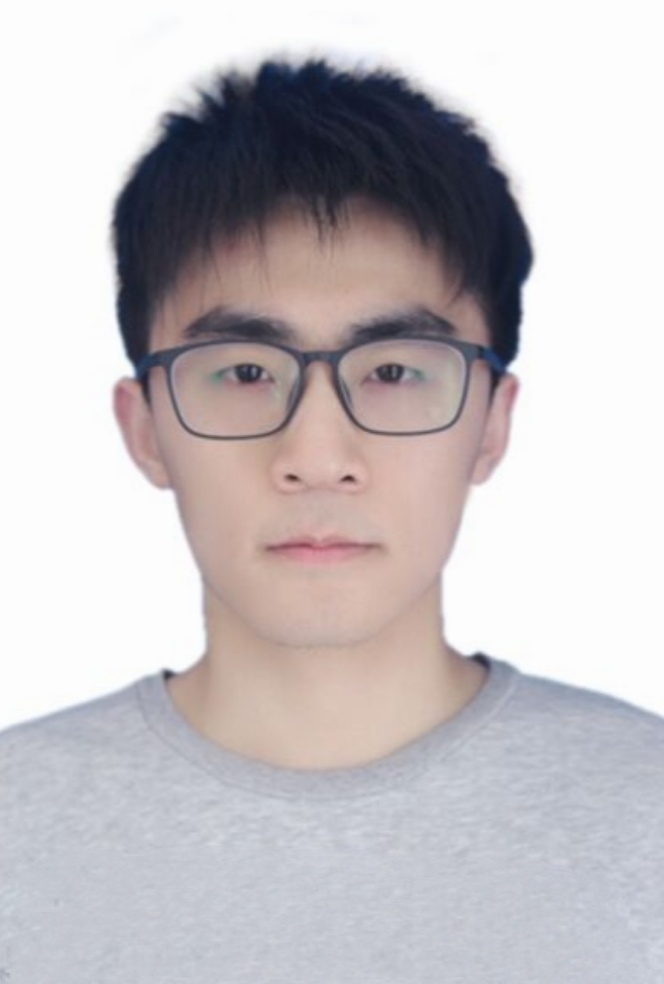}
}]{Yifeng Xiong}
is currently pursuing the master degree with Huazhong University of Science and Technology, Wuhan, China.  He received the B.S. degree in computer science from Huazhong University of Science and Technology, Wuhan, China, in 2020. His research interests include adversarial learning, machine learning, computer vision and natural language processing.
\end{IEEEbiography}

\begin{IEEEbiography}[{\includegraphics[width=1in,height=1.25in,clip,keepaspectratio]
{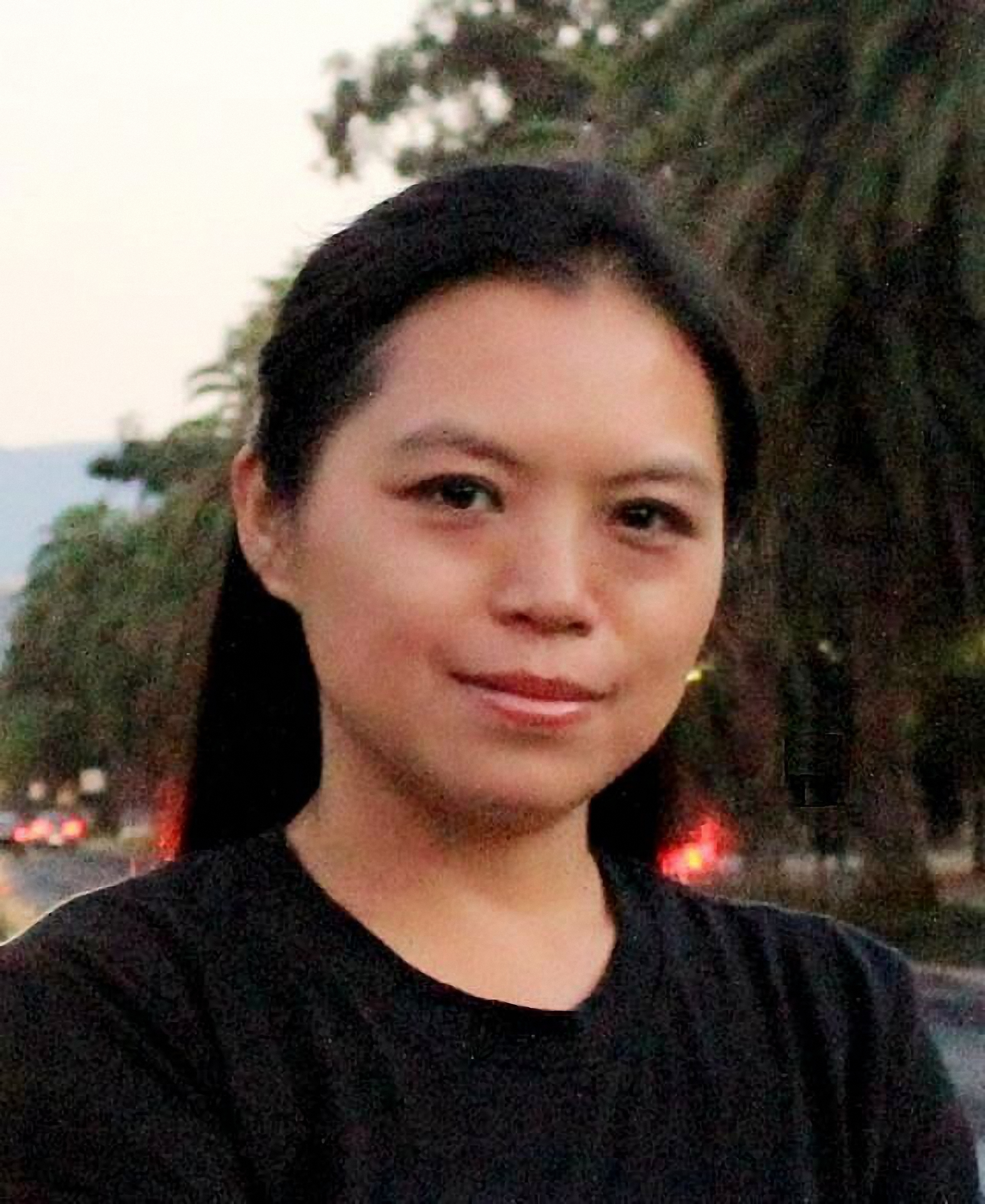}
}]{Kun He} (SM18) 
is currently a Professor in School of Computer Science and Technology, Huazhong University of Science and Technology, Wuhan, P.R. China; and a Mary Shepard B. Upson Visiting Professor for the 2016-2017 Academic year in Engineering, Cornell University NY, USA. She received 
the Ph.D. degree in system engineering from Huazhong University of Science \& Technology, 
in 2006. Her research interests include adversarial learning, machine learning, social network analysis, and combinatorial optimization.
\end{IEEEbiography}
\vfill

\appendix{The supplementary material for Semantic Adversarial Attacks on Face Recognition through Significant Attributes.}

\section{Semantic Adversarial Attack (SAA-StarGAN)}\label{sec11}
\subsection{Cosine Similarity (CS)}
\begin{algorithm}[h] 
\caption{Determine the most significant attributes by CS}
\label{alg:1}
\begin{algorithmic}[1]
\Require{Original image \(x\), attributes \( \{a\}\), 
     generator \(G\) of StarGAN, target face verification model \(TFV\)} 
\Ensure{The most significant attributes $C$}
\For {each $a_i$ in attributes $\{a\}$} \label{algorw1}  
    \State {change $a_i$ using $G$}
    \State{\(x^*_{a_{i}}\) $\gets$ $G$($x$, $a_i$)}
    \State{$f_{x^*} \gets TFV$ $(x)$, $f_{x^*_{a_i}} \gets TFV (x^*_{a_{i}})$} \label{algorw4}
    \State{$S_{a_i}$ $\gets CS(f_x, f_{x^*_{a_i}})$  } 
\EndFor
\State{$C$ $\gets$ Sort $\{a\}$ according to each $S_{a_i}$ in the ascending order}
\State{Output the most significant attributes $C = (c_1, c_2, c_3,...,c_K)$} 
\end{algorithmic}
\end{algorithm}
\subsection{Probability Score (PS)}
\begin{algorithm}[h] 
\caption{Determine the most significant attributes by PS}
\label{alg:2}
\begin{algorithmic}[1]
\Require{Original image \(x\), attributes \(\{a\}\), 
         generator \(G\) of StarGAN, attribute prediction model \(f\) } 
\Ensure{ The most significant attributes \(\{C\}\)}
\State{Use attribute prediction model to predict attributes}
\For{ each $a'_i$ in attributes $\{a\}$}  
    \State{$P_{a'_i}(x)$ $\gets$ $f(x)$}
    \State {Change $a'_i$ using $G$}
    \State{\(x^*_{a'_{i}}\) $\gets$ $G$($x$, $a'_i$)}
    \State{$P_{a'_i}(x^*_{a'_i})$ $\gets$ $f$( $x^*_{a'_i}$) \Comment{Compute the probability for each attribute}}
    \State{Compute $\Delta P_{a'_i }$} 
\EndFor
\State{$C$ $\gets$ Sort $\{a\}$ using $\Delta P_{a'_i }$ in descending order}
\State{Output the most significant attributes $C = (c_1, c_2, c_3,...,c_K) $ }
\end{algorithmic}
\end{algorithm}
\subsection{SAA-StarGAN in Black-Box Setting}
\begin{algorithm}[t] 
\caption{The SAA-StarGAN for Black-Box Attack}
\label{black}
\begin{algorithmic}[1]
\Require{Original image \(x\), original attribute $c_{ori}$, most significant attributes  \(\{c\}\), encoder \(G_E\), decoder \(G_D\),  target face verification model \(TFV\), target image \(x_{tgt}\), $L_2$ distance function $dist$, threshold\_sim $th$ } 
\Ensure{Advesarial example \(x_{adv}\)}
\State{$f^*_ 1$ $\gets$ $G_E$ ($x$, $c_{ori}$)}
\State{$\gamma$ $\gets$ create array in range [0,1]} 
\For { $c_i$ in most significant attributes $C$} 
    \State{$f^*_{c_ i}$ $\gets$ $G_E$ ($x$, $c_i$)}
       \For  { $\gamma_i$ in $\gamma$}
            \State {$f^*_ {\gamma_i}$ = $\gamma_i$ $\cdotp$ $f^*_ 1$ + (1- $\gamma_i$) $\cdotp$ $f^*_{c_ i}$ }
            \State {$x^*_{\gamma_i}$ = $G_D$ ($f^*_ {\gamma_i}$)}
            \State {$d_{\gamma_i}$ $\gets$ $dist$ ($TFV(x_{tgt}, x^*_{\gamma_i})$}
             \State {$score.insert(d_{\gamma_i}, \gamma_i $ )}
            \EndFor
            \If{impersonation attack } 
                \State {$\gamma_{opt}$ = $argmax_{\gamma}$ ($score$)}
            \ElsIf{dodging attack}
                \State  {$\gamma_{opt}$ = $argmin_{\gamma}$ ($score$)}
            \EndIf
             
            \State{Substitute $\gamma_{opt}$ in $f^*$}
            \State{$x^*$ = $G_D$ ($f^*$)}
            \If{$sim (x, x^*)\leq  th$} \label{algorw18}
                \State Return None
            \EndIf
            \If{$dist(TFV(x^*,x_{tgt})\leq T)$}
                \State $x_{adv} \gets x^* $
                \State{break}
            \Else
                \State $x \gets x^*$
                \State $c_i \gets$ add the next attribute from $C$ 
            \EndIf
            \If{$dist(TFV(x^*,x)\geq T)$}
                \State $x_{adv} \gets x^* $
                \State{break}
            \Else
                \State $x \gets x^*$
                \State $c_i \gets$ add the next attribute from $C$ 
            \EndIf \label{algorw34}
        \EndFor
\end{algorithmic}
\end{algorithm}
\section{Evaluations on Cosine Similarity} 
\begin{figure*}[t]
\centering
\includegraphics[width=0.9\textwidth]{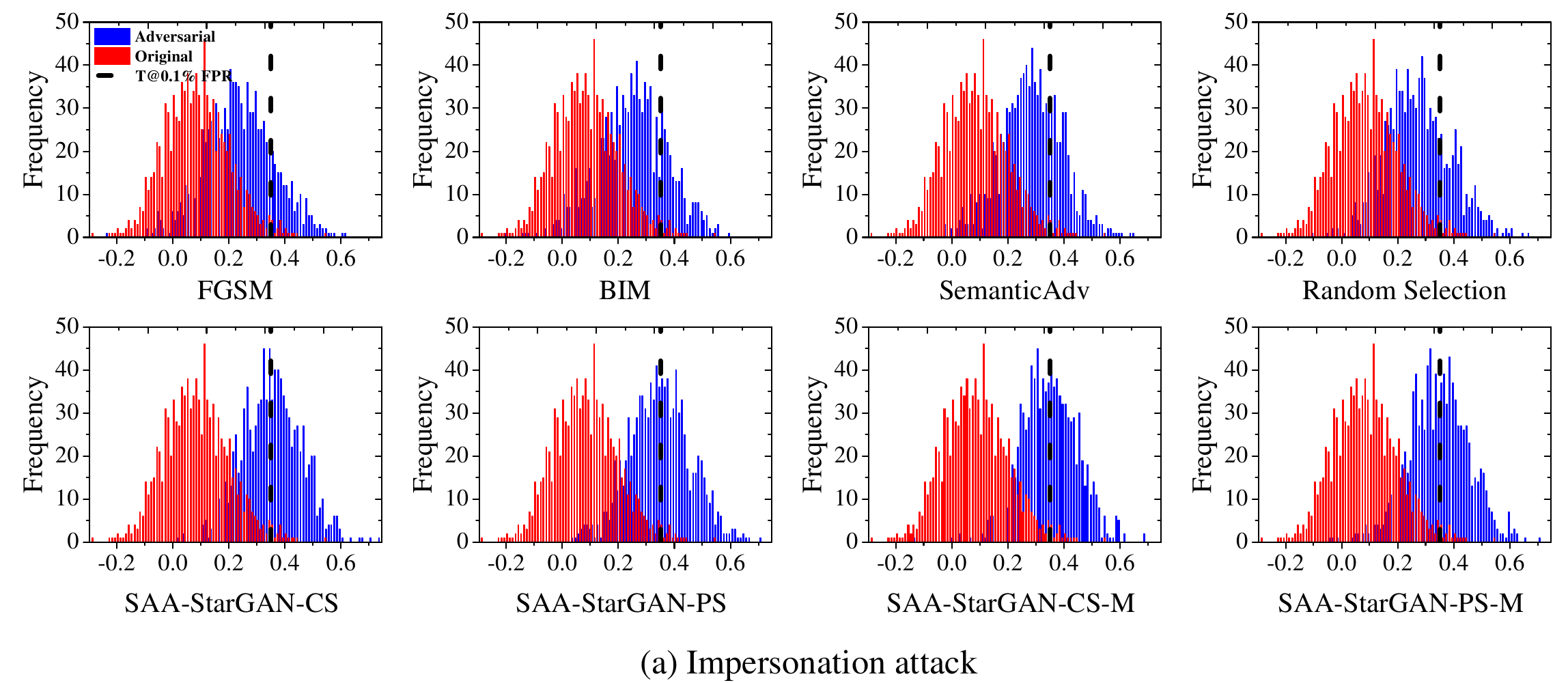}
\includegraphics[width=0.9\textwidth]{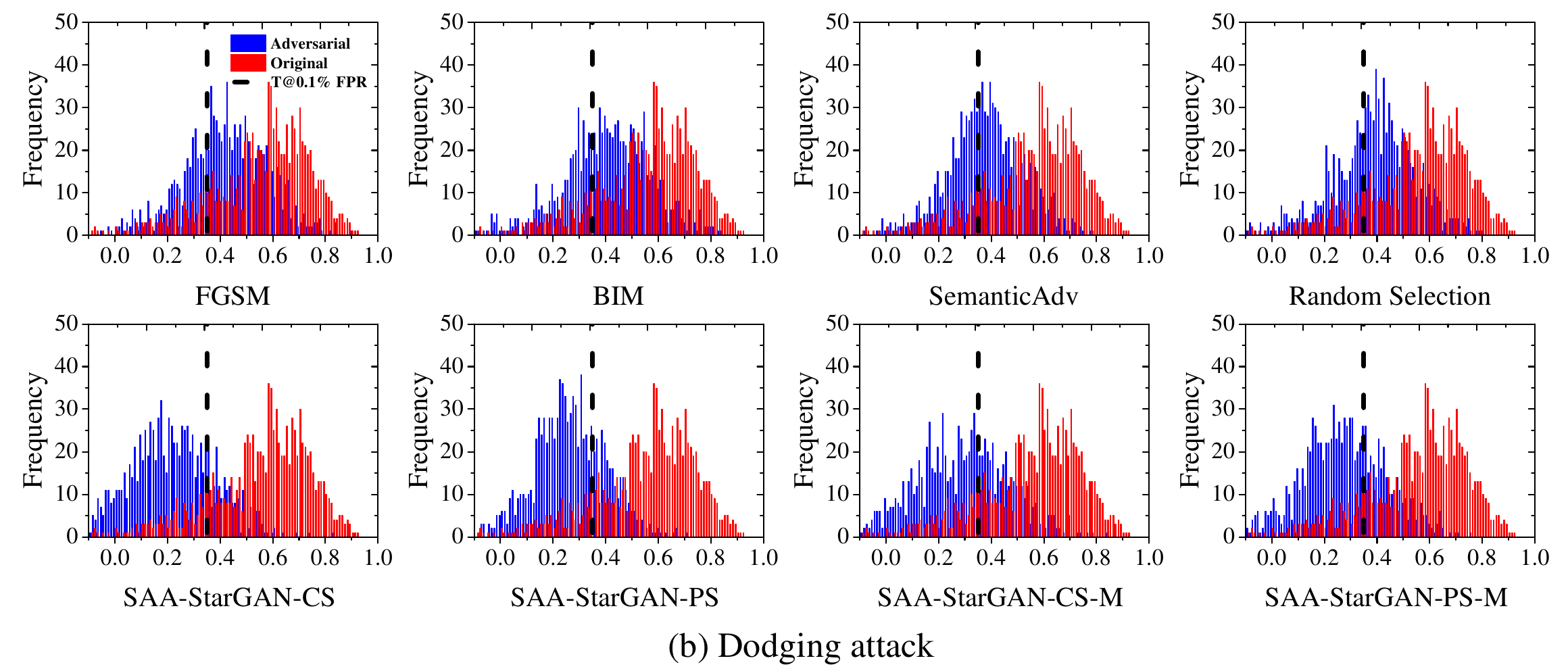}
\caption{Shift in cosine similarity scores for SphereFace before and after adversarial attacks generated by SAA-StarGAN and the baselines on FaceNet. 
The shift of SAA-StarGAN is much more distinct as compared with the baselines.
}
\label{Cosine}
\end{figure*}

To emphasize and clarify the effectiveness of the proposed SAA-StarGAN method in improving the transferability against different models. We select 1000 adversarial face images crafted on FaceNet from each attack method against a SphereFace model under an impersonation attack.
For the dodging attack, we choose 1000 of the generated adversarial face images on FaceNet with the original images. We measure the cosine similarity scores before and after the attack to see the improvement percentage for each attack method against a SphereFace model. In Fig. \ref{Cosine}, we observe that most scores between the generated face and target faces of the SAA-StarGAN method fall above $T_s$ @ 0.1 \% FPR, demonstrating that the generated face images can be falsely accepted by 53, 50, 48.7, and 50.6 \% in an impersonation attack. In contrast to the baseline methods, a few scores fall above $T_s$. For the dodging attack, the scores fall below $T_s$ @ 0.1 \% FPR by 77.2, 71.8, 61.4, and 66.7 \%, demonstrating that the SphereFace model can falsely reject the image pairs. We conclude that the SAA-StarGAN significantly outperforms the baseline attacks to improve the transferability of black-box models. 

\end{document}